\documentclass[11pt]{article}
\usepackage[numbers]{natbib}
\usepackage{macros/packages}
\usepackage{macros/editing-macros}
\usepackage{macros/formatting}
\usepackage{macros/statistics-macros}
\usepackage[utf8]{inputenc} 
\usepackage[T1]{fontenc}    
\usepackage{hyperref}       
\usepackage{url}            
\usepackage{booktabs}       
\usepackage{amsfonts}       
\usepackage{listings}  
\usepackage{verbatim}  
\usepackage{nicefrac}       
\usepackage{microtype}      
\usepackage{xcolor}         
\usepackage{wrapfig}
\usepackage{subcaption}
\usepackage{float}
\usepackage{tikz}
\usepackage{algorithm}
\usepackage{algorithmic}

\usetikzlibrary{%
  calc,
  fit,
  arrows,
  arrows.meta,
  positioning,
  decorations.pathreplacing,
  decorations.shapes,
}
\usepackage[textsize=tiny]{todonotes}
\theoremstyle{definition}

\usepackage{cleveref}
\usepackage{subcaption}

\usepackage{ulem}

\renewcommand{\emph}[1]{\textit{#1}}

\begin{document}

\abovedisplayskip=8pt plus0pt minus3pt
\belowdisplayskip=8pt plus0pt minus3pt


\begin{center}
  {\huge LLM Generated Persona is a Promise with a Catch} \\
  \vspace{.5cm} {Ang Li ~~~~ Haozhe Chen ~~~~  Hongseok Namkoong ~~~~ Tianyi Peng} \\
  \vspace{.2cm}
  {\large Columbia University} \\
       \vspace{.2cm}
   \texttt{\{al4263, haozhe.chen, hongseok.namkoong,  tianyi.peng\}@columbia.edu}
\end{center}


\begin{abstract}%
  The use of large language models (LLMs) to simulate human behavior has gained significant attention, particularly through personas that approximate individual characteristics. Persona-based simulations hold promise for transforming disciplines that rely on population-level feedback, including social science, economic analysis, marketing research, and business operations. Traditional methods to collect realistic persona data face significant challenges. They are prohibitively expensive and logistically challenging due to privacy constraints, and often fail to capture multi-dimensional attributes, particularly subjective qualities. Consequently, synthetic persona generation with LLMs offers a scalable, cost-effective alternative. However, current approaches rely on ad hoc and heuristic generation techniques that do not guarantee methodological rigor or simulation precision, resulting in systematic biases in downstream tasks. Through extensive large-scale experiments including presidential election forecasts and general opinion surveys of the U.S. population, we reveal that these biases can lead to significant deviations from real-world outcomes. Our findings underscore the need to develop a rigorous science of persona generation and outline the methodological innovations, organizational and institutional support, and empirical foundations required to enhance the reliability and scalability of LLM-driven persona simulations. To support further research and development in this area, we have open-sourced approximately one million generated personas, available for public access and analysis at \href{https://huggingface.co/datasets/Tianyi-Lab/Personas}{Tianyi-Lab/Personas}
\end{abstract}

\section{Introduction} 
\begin{figure*}[!t]
\centering\includegraphics[width=\textwidth]{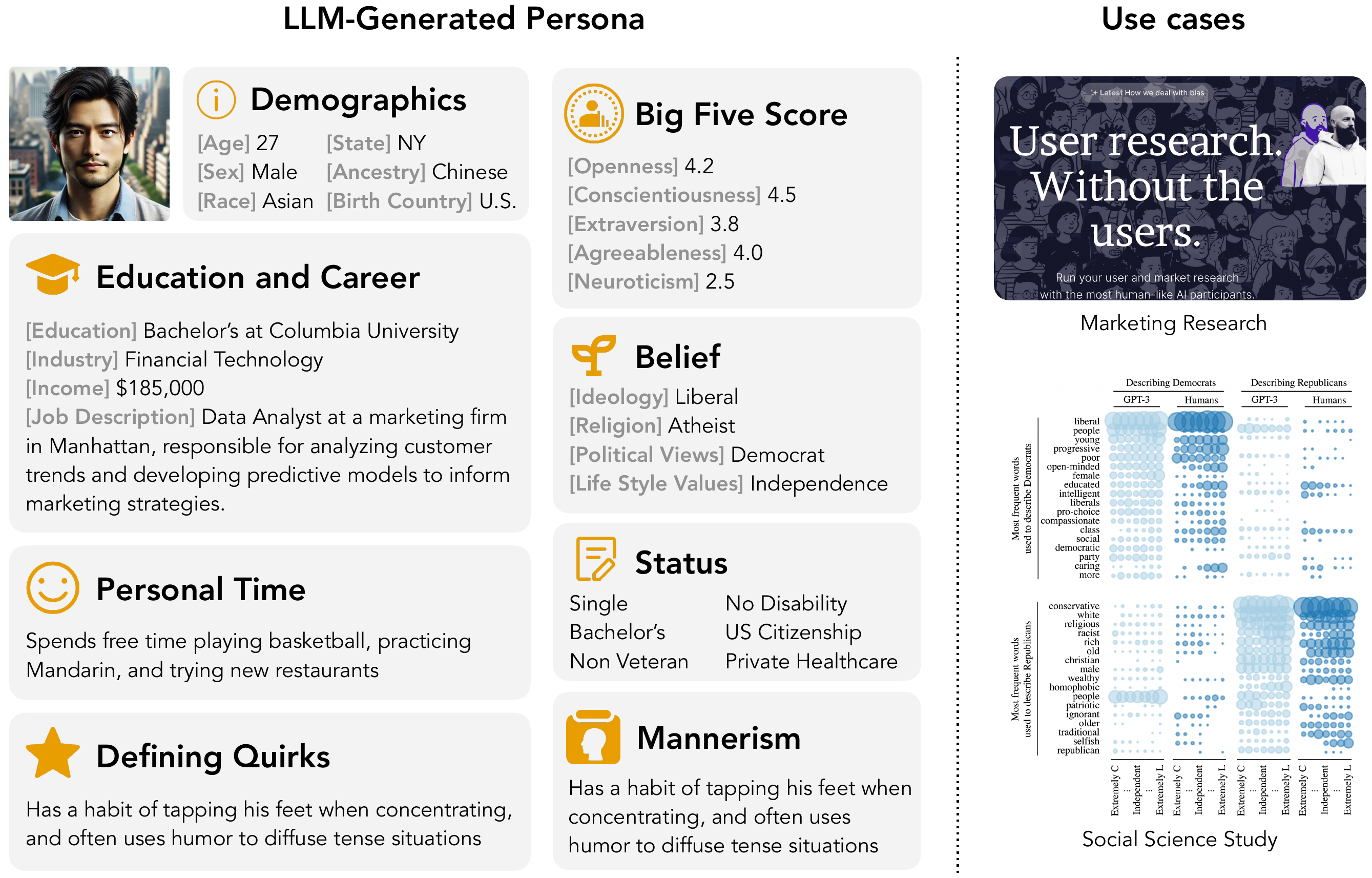}
    \caption{Left: An example of a LLM generated persona. Right: Applications of personas in the real world.}
    \label{fig:persona-examples}
\end{figure*}

Recent advances in LLM-driven simulations have enabled the creation of synthetic agents that approximate real-world populations' behaviors at scale, potentially transforming various fields such as social science \cite{manning2024automated}, political science \cite{argyle2023out}, economics \cite{horton2023large}, marketing research \cite{sarstedt2024using}, clinical psychology \cite{wang2024patient}, entertainment \cite{shao2023character} and business applications such as recommendation systems and web testing \cite{tseng2024two}. A central component of these simulations is the use of \textbf{personas} as shown in \Cref{fig:persona-examples}—digital representations of individuals characterized by demographic, psychographic, and behavioral attributes. When provided with persona information, LLMs can generate responses and simulate decision-making processes that aim to reflect those of real individuals \citep{park2024generative}, serving as ``silicon samples'' to complement ``human samples'' for opinion testing~\cite{argyle2023out}. 

While ``silicon samples'' hold the potential to revolutionize societal-scale experimentation, a critical challenge remains: the systematic and reliable generation of persona profiles for targeted populations.  Although large-scale collection of human data can maximize downstream simulation fidelity \citep{park2024generative}, gathering such comprehensive data is far from trivial. It requires extensive, resource-intensive surveys and coordinated efforts that demand substantial financial investment, time, and operational support, while adhering to strict privacy regulations and ethical guidelines. On the other hand, existing large-scale real-world datasets (e.g., the U.S. Census \cite{uscensus2024}) primarily contain marginal demographic information without capturing the joint distribution of multi-dimensional attributes. This fragmentation makes it impossible to reconstruct realistic, integrated personas that accurately mirror the complexity of real-world populations. Furthermore, these datasets often omit subjective attributes—such as lifestyle preferences or nuanced belief systems—due to privacy concerns or measurement limitations, despite their critical role in shaping individual opinions and values.

Meanwhile, LLM itself presents a viable solution to the above challenges by generating persona profiles directly in a cost-effective, efficient, and seemingly realistic manner.
This nascent direction has received much attention  recently from both academia and industry \cite{frohling2024personas,castricato2024personareproducibletestbedpluralistic, ge2024scalingsyntheticdatacreation,schuller2024generating,syntheticusers,civicsync}, where LLM-generated personas have been used to conduct surveys, marketing research, or even societal-scale simulations.  
Prior work has primarily focused on methodologies for scaling up the number of diverse personas, but there are no rigorous evaluations of their performance in different downstream applications or whether they faithfully capture a specific population's opinion at scale.

\begin{figure*}[!t]
\centering\includegraphics[width=\textwidth]{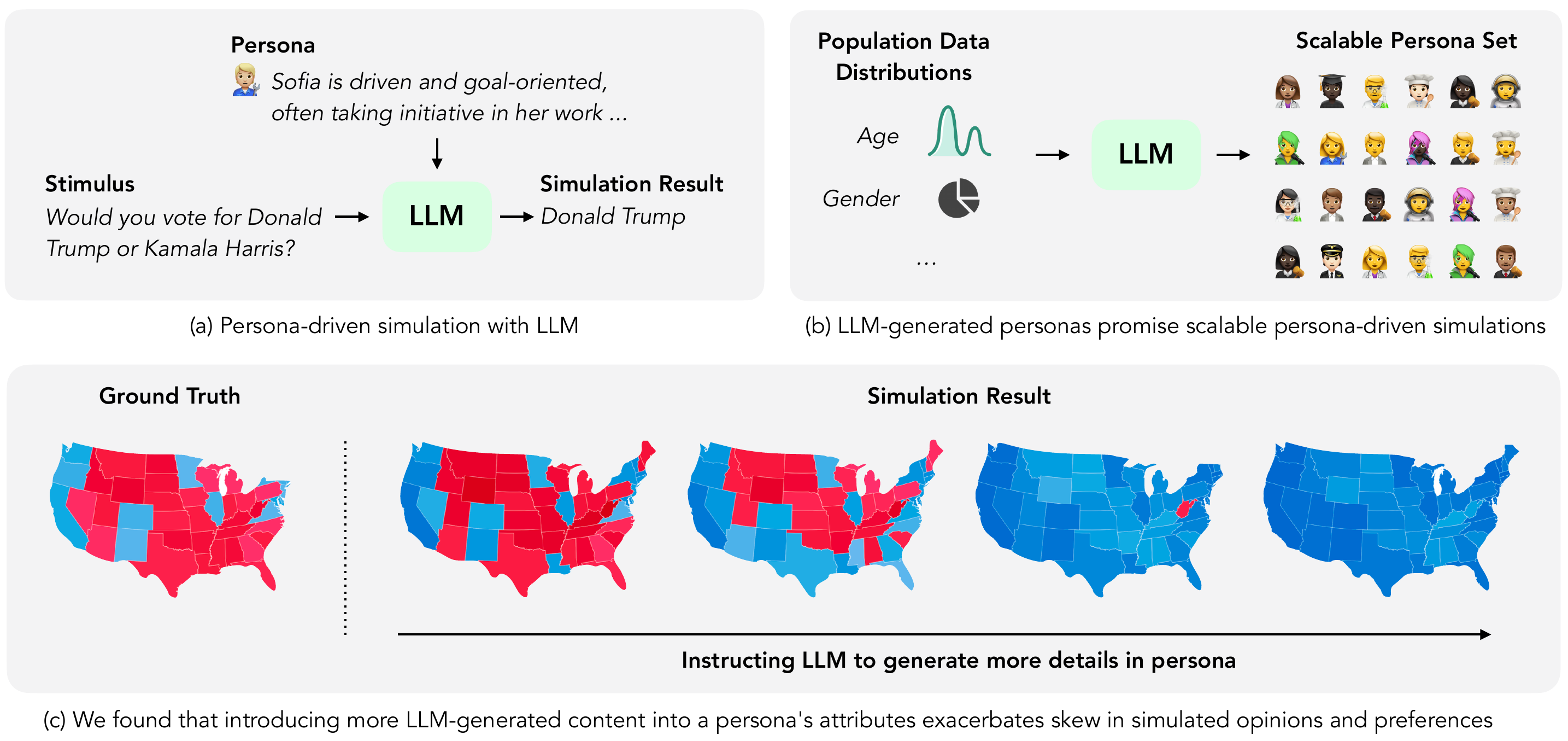}
    \caption{(a) Persona-driven simulation enables simulating human behaviors with LLMs. (b) Using LLM to generate personas promises scalable simulation of diverse population's behaviors.  (c) We caution that improper usage of LLMs as persona generators may lead to homogeneous results.}
    \label{fig:persona-teaser}
\end{figure*}

Given the huge potential and interests of LLM-based persona generation and the lack of a scientific community that studies this problem, we argue that \textbf{a science of persona generation needs to be developed to fully realize the potential of LLM persona simulation.} Specifically, we observe that the current scalable persona generating methods are significantly biased and non-representative of the real-world distribution. \textit{One example is that in a 2024 presidential election simulation, results generated by a specific type of LLM-based synthetic personas predict a Democratic sweep across all U.S. states} (see \Cref{fig:persona-teaser}). Since these generated personas can be widely used in applications ranging from opinion simulation to product testing, \textit{their inherent biases can lead to harmful consequences, including skewed public decision-making, reinforcement of discrimination and stereotypes, and potential harm to minority groups.}

While collecting better persona data is vital and nontrivial, we believe that both data collection and persona generation and simulation methods are complementary rather than mutually exclusive. Regardless of the dataset's size, generation methods are necessary to capture the complex, subjective aspects that traditional surveys often miss.

Therefore, we see an urgent need and call for a concerted effort from the AI community and interdisciplinary collaborations across social sciences to develop reliable methodologies, robust large-scale benchmarks, and broad community support to establish persona generation and silicon-sample simulation as a rigorous field. 

This paper is organized as follows:
\begin{enumerate}
    \item \textbf{Systematization of Persona Generation Methods.} In Section \ref{sec:persona-generation}, we categorize and systematize existing persona generation approaches, as summarized in \Cref{fig:overview-persona}. This systematization lays the foundation for subsequent experimental evaluation and provides a structured framework for advancing the methodological rigor of persona generation.
 
    \item \textbf{Current Persona Generation Practice Falls Short:} In Section \ref{sec:persona-results}, we present an extensive experimental evaluation of current persona generation practices. Using three commonly utilized persona types, we generate around 1,000,000 personas with six open-source LLMs and assess them across 500+ questions spanning diverse domains. The results demonstrate a substantial bias, amplified by the LLM-generated persona content. These findings critically undermine the validity of using LLMs as "silicon samples" and raise serious concerns about applying LLM-generated personas in social science research and business decision-making without proper caution.

    \item \textbf{Analysis of Persona Profiles:} Section \ref{sec:bias-in-profiles} provides a semantic analysis of the LLM-generated persona profiles.  The analysis reveals the sources of the observed bias and suggests potential mitigation strategies.

    \item \textbf{A Path Towards Rigorous Persona Generation:} In Section \ref{sec:rigorous-persona-vision}, we outline key directions for establishing a more scientific and systematic approach to persona generation, including: (i) developing refined frameworks for identifying essential persona attributes, (ii) building theoretical and empirical foundations for persona profile calibration, (iii) creating large-scale datasets and benchmarks to validate and refine persona generation methods, and (iv) advocating for interdisciplinary research and careful application-oriented consideration of risks and biases in persona generation and simulation.
\end{enumerate}
Overall, we advocate for increased attention from the broader research community to realize the transformative potential of widespread LLM silicon sample adoption, enabling rapid, cost-effective, safe, and human-centric tests and innovation. As a first step toward this goal, we open-source all generated personas in our experiments to support and accelerate community efforts in this direction.

\paragraph{A study of LLM persona generation bias rather than traditional LLM bias.}  While biases in LLMs have been extensively studied in the literature \cite{gallegos2024bias, ferrara2023should, feng2023pretraining, parrish2021bbq, bai2024measuring, lu2020gender, li2020unqovering}, most prior work focuses on biases arising from \textbf{LLM simulation}.  In contrast, we highlight a striking and underexplored source of bias introduced by \textbf{LLM persona generation} itself. Specifically, prior work in LLM bias has largely taken the persona generation process for granted, focusing instead on simulation outputs. In contrast, our study explicitly dissects the persona generation step, revealing that even if the simulation appears unbiased, the underlying generated personas may carry significant, unexamined biases that undermine their validity. Our findings indicate that this area requires significantly more attention to ensure fairness and representativeness in generated personas.
\section{Persona-Driven Opinion Simulation}
\label{sec:persona-generation}

To highlight the shortcomings of current persona generation methods, we first establish the foundation for large-scale simulation experiments by implementing various commonly used persona generation strategies and simulation methods, drawing from both prior literature and community practices.

\subsection{Persona Generation}

We build on prior literature to construct three distinct types of personas: \textbf{Meta Personas}, \textbf{Tabular Personas}, and \textbf{Descriptive Personas}. 

\paragraph{Meta Persona.} Census data provides a foundational source for generating realistic personas due to its rich demographic information that reflects real-world population distributions. However, census data typically presents marginal distributions (e.g., percentages of different age groups and genders in New York), which limits its capacity for capturing joint probabilities. This limitation can lead to incongruous combinations of attributes (e.g., a person earning \$500k annually but unable to afford health insurance). To address this, we adopt methods from \citet{chang2024llmsgeneratestructurallyrealistic} and \citet{argyle2023out}, leveraging data with joint distributions over key attributes, such as Age, Sex, Race, and State, provided by the U.S. Census Bureau. We term these structured, sampled personas \textbf{Meta Personas} (No LLM is involved in generating them). While these personas may lack diversity, they serve as a robust foundation for generating more varied and realistic personas.

\begin{figure*}[!t]
    \centering      \includegraphics[width=\textwidth]{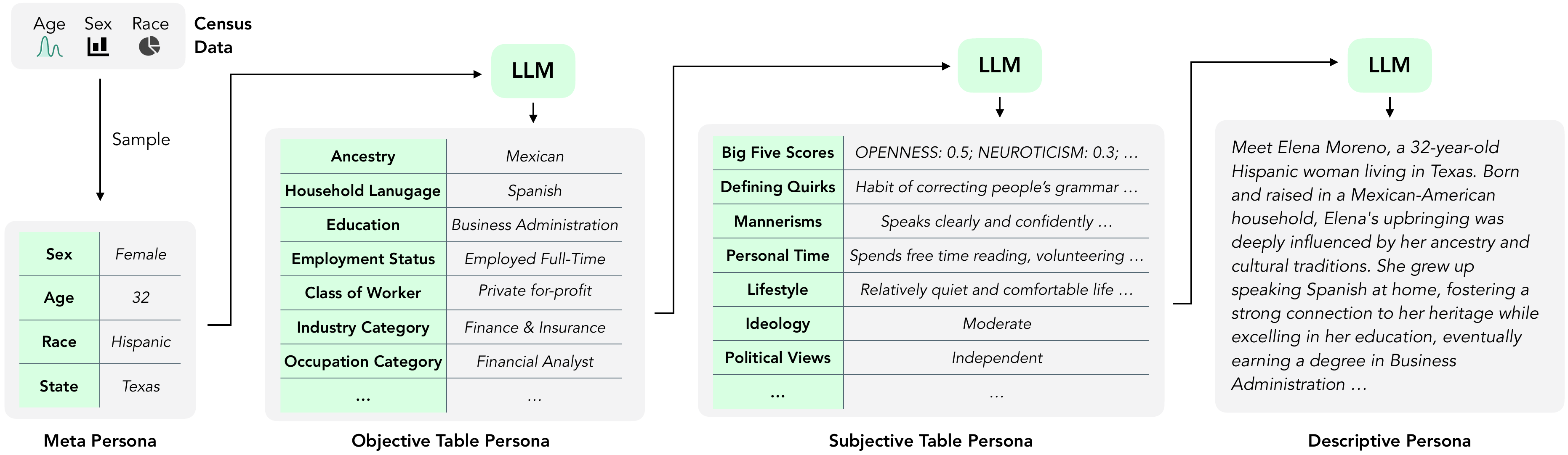}
    \caption{We categorize existing persona generation approaches into four tiers. Each tier adds more information generated by LLMs in generated personas to the previous tier.}
\label{fig:overview-persona}
\end{figure*}

\paragraph{Tabular Personas.} To enhance diversity and better capture nuanced differences for opinion simulation, we extend the information encoded in \textbf{Meta Personas} using structured templates. Following the proposed method of \citet{castricato2024personareproducibletestbedpluralistic}, we employ LLMs to fill in additional attributes within these templates, conditioned on the generated meta personas. We propose two variants:

1. \textit{Objective Tabular Personas}: These personas expand \textbf{Meta Personas} by adding attributes that can be objectively measured (e.g., Income, Education Level, Occupation, and Industry). We align these entries with the Census Bureau's predefined categories (detailed in the appendix), which ensures consistency and mitigates bias. For example, occupations are drawn from five predefined categories rather than generated freely by the LLM.

2. \textit{Subjective Tabular Personas}: These personas extend \textit{Objective Tabular Personas} by adding subjective or less standardized attributes (e.g., Political Affiliation, Leisure Preferences). Since these attributes lack comprehensive predefined categories, we prompt the LLM to generate open-ended responses, guided by templates designed to maximize diversity while preserving realism.

\paragraph{Descriptive Personas.} The most flexible approach, \textbf{Descriptive Personas}, involves directly prompting the LLM to create freeform personas conditioned on \textbf{Meta Personas}. This method allows the generation of highly detailed and unconstrained personas by asking the LLM to expand beyond structured attributes into narrative-style descriptions. While this approach provides maximum freedom, it requires careful prompting and post-processing to ensure the plausibility and coherence of the generated personas. 

Progressing from \textbf{Meta Personas} to \textbf{Tabular Personas} to \textbf{Descriptive Personas} entails a growing content generated by LLMs for diversity. This means that the persona types from left to right in \Cref{fig:overview-persona} gradually introduce richer, more varied attributes and open-ended descriptions, which theoretically can yield broader and more diverse opinion distributions. However, as we show later, \textbf{increasing the amount of LLM-generated persona content can exacerbate biases in the simulation results in various domains.} 

To approximate opinion distributions and capture diversity at the U.S. national level, we first sample 1,000 personas per state using the corresponding joint distribution. We then generate 1,000 augmented personas for each of the persona types (\textbf{Meta}, \textbf{Tabular}, and \textbf{Descriptive}). During the opinion simulation, we present the language model with a short survey: a question accompanied by multiple choices, a persona description, and a template prompting the model to select the most suitable response based on the persona's attributes. We explore several prompt variants that encourage neutrality and mitigate unwanted biases or stereotypes, but observe no substantial performance differences across these variants. The specific prompts and templates can be found in \Cref{prompts}

\subsection{Simulating Opinions with LLMs}

We treat LLMs as impartial world models and simulators. Building on recent work such as \citet{park2024generative}, we hypothesize that, when conditioned on realistic and adequate information, LLMs can truthfully simulate a persona's opinion, effectively serving as digital twins. To simulate each persona's opinion, we first provide high-level instructions guiding the LLM to remain unbiased and objective while faithfully reflecting the persona’s attributes. Next, we input a persona along with a question formatted as a multiple-choice query, instructing the LLM to directly select one of the provided answers. The final prompt, includes the simulation instruction, persona description, and the simulation question, is fed once into the language model. Additional simulation details can be found in \Cref{Additional Simulation Details}.
\section{Can Personas Simulate the Society?}
\label{sec:persona-results}

\begin{figure}[!t]
    \centering
    \begin{subfigure}[t]{1.0\textwidth}
        \centering
        \includegraphics[width=\textwidth]{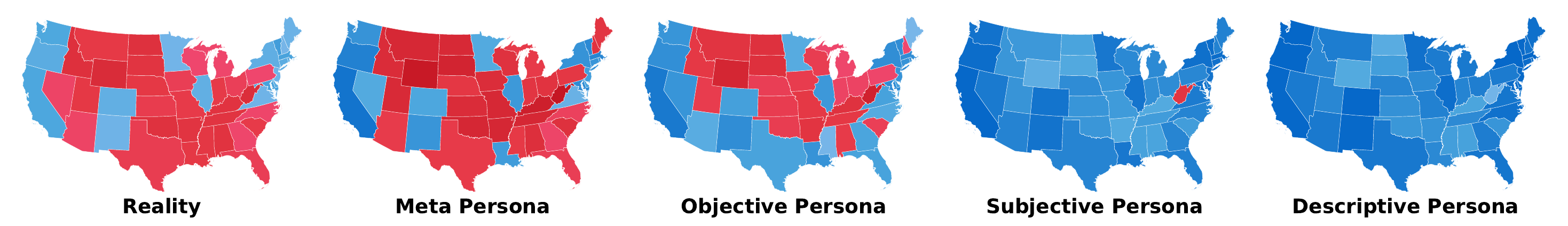}
        \caption{2024 Election}
        \label{fig:2024}
    \end{subfigure}
    
    \vspace{1em} 
    
    \begin{subfigure}[t]{1.0\textwidth}
        \centering
        \includegraphics[width=\textwidth]{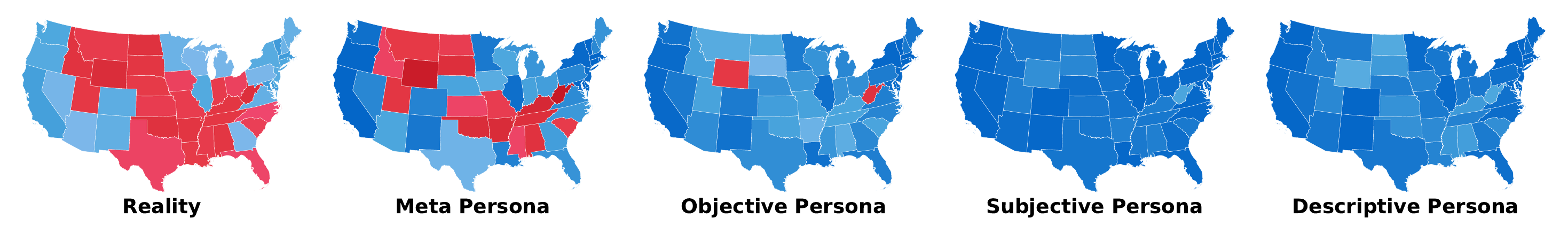}
        \caption{2020 Election}
        \label{fig:2020}
    \end{subfigure}
    
    \vspace{1em}
    
    \begin{subfigure}[t]{1.0\textwidth}
        \centering
        \includegraphics[width=\textwidth]{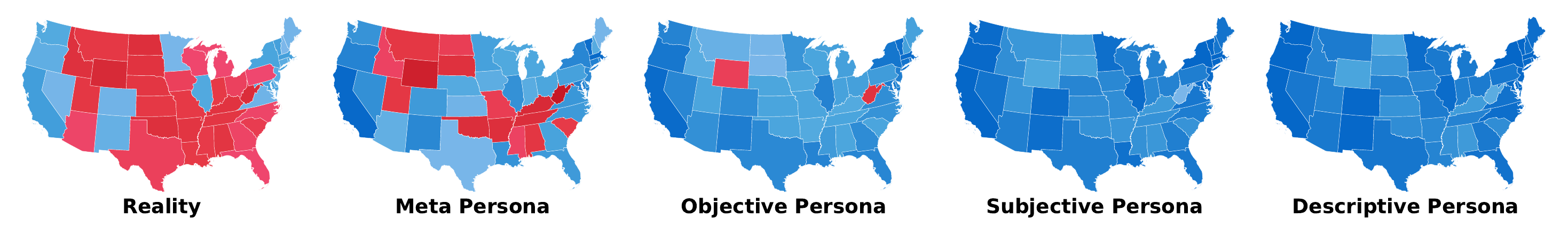}
        \caption{2016 Election}
        \label{fig:2016}
    \end{subfigure}
    
    \caption{Persona-based simulations of elections 2016, 2020, and 2024.}
    \label{fig:overview}
\end{figure}

We now present empirical findings from two main simulation scenarios: (1) U.S. elections from 2016, 2020, and 2024, and (2) general opinion surveys drawn from OpinionQA \citep{santurkar2023whose}. Our analysis focuses on how relying on different persona types (with varying amounts of LLM-generated content) affects alignment with real-world data.

\begin{figure*}[!t]
    \centering  
    \includegraphics[width=\textwidth]{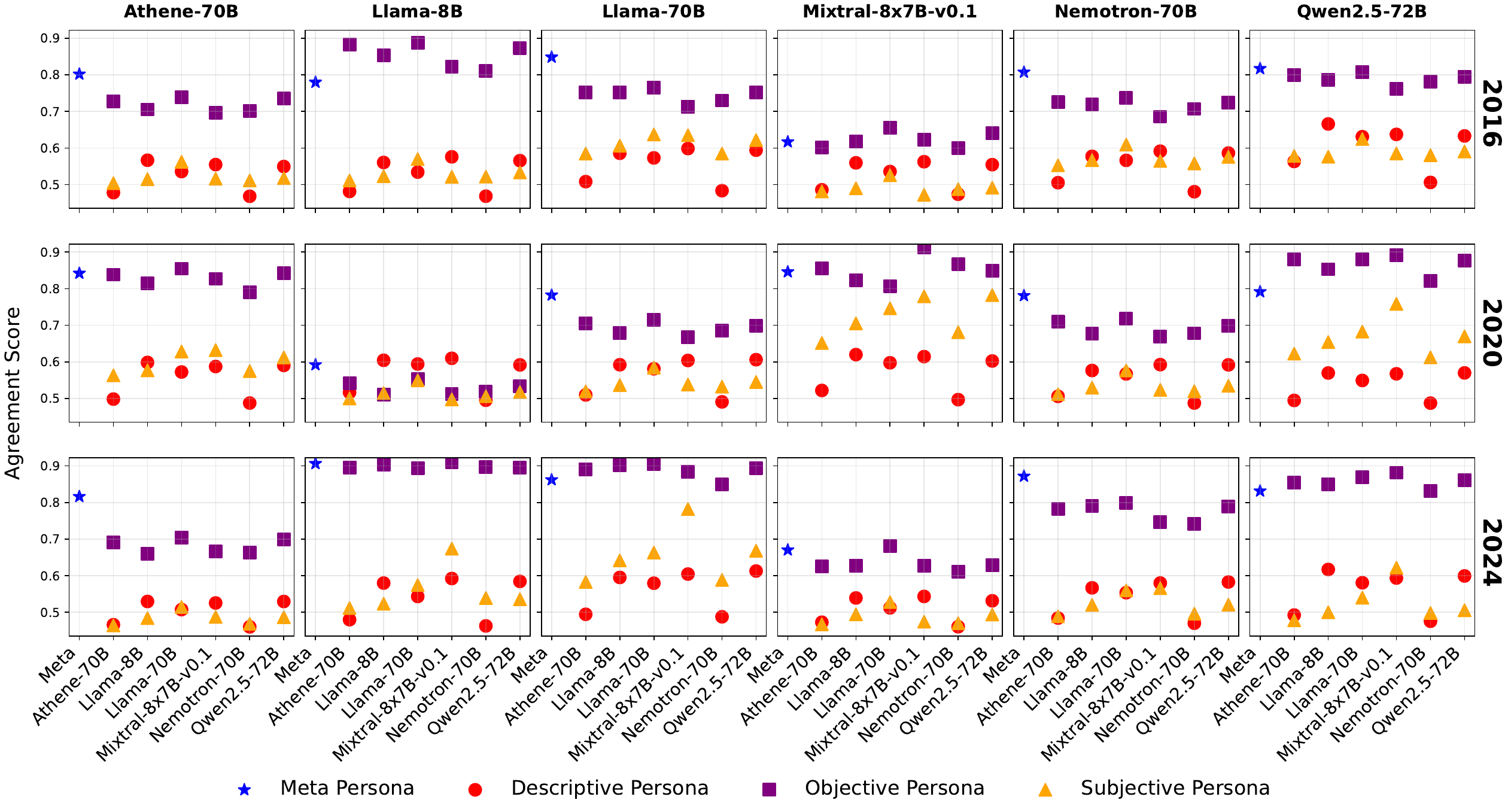}
    \caption{Alignment scores for cross-model simulation. Each column represents a simulation model, while the x-axis within each column corresponds to the persona generation model. The ``meta'' point is singular as it relies on sampling rather than generation.}
    \label{fig:alignment scores}
\end{figure*}

\subsection{A Case Study with US Presidential Election Voting}

Motivated by prior work in political and social opinion simulations \cite{argyle2023out}, we explore how our persona-based approach can serve as a viable methodology for modeling society-level electoral outcomes. Given that most large language models (LLMs) are trained on data collected prior to the 2024 U.S. election, we first test on the 2024 presidential election to avoid potential confounds from post-2024 information.

We plot the map of each state in blue and red with different color scales to indicate the support rate. In \Cref{fig:2024}, we show one particular experimentation where the persona model and the simulation model are both Llama 3.1 70B. Our results reveal that with increasing LLM-generated persona attributes, the simulation results tend to deviate more from real-world outcomes. Specifically, \textbf{Meta Personas} yield simulated voting distributions closest to reality data, whereas \textbf{Generative Personas} produce results most divergent from reality. Moreover, we observe that as the level of LLM-generated persona information rises, the simulated electorate shifts progressively toward left-leaning stances, ultimately culminating in a surprising scenario where every state appears to vote for the Democratic candidate.

To investigate whether historical election data—presumably included in the training sets—reduces model bias, we conduct similar simulations for the 2016 and 2020 election cycles. One might expect that since these events are part of the training data, the models would “remember” historical facts and produce more accurate electoral distributions. Contrary to this expectation, we observe a similar pattern of leftward drift, suggesting that awareness of past elections alone does not sufficiently counterbalance the influence of LLM-generated content.

\paragraph{Quantitative evaluation.} To more rigorously compare our simulated results with ground truth, we adopt the alignment metric from \citet{santurkar2023whose}, using $1 - W(\hat{p}, p)$ as the alignment score. Here, $W(\hat{p}, p)$ denotes the Wasserstein distance between the simulated support rate (\(\hat{p}\)) and the actual voting rate (\(p\)), with higher alignment scores indicating results closer to real-world outcomes. Additionally, we perform \textit{cross-simulation}, in which each LLM-generated persona set is simulated across all other language models. Our comprehensive assessment \Cref{fig:alignment scores} shows that this type of bias is not model-specific but rather universal in persona-driven simulations. 

\begin{figure*}[!t]
    \centering  
    \includegraphics[width=\textwidth]{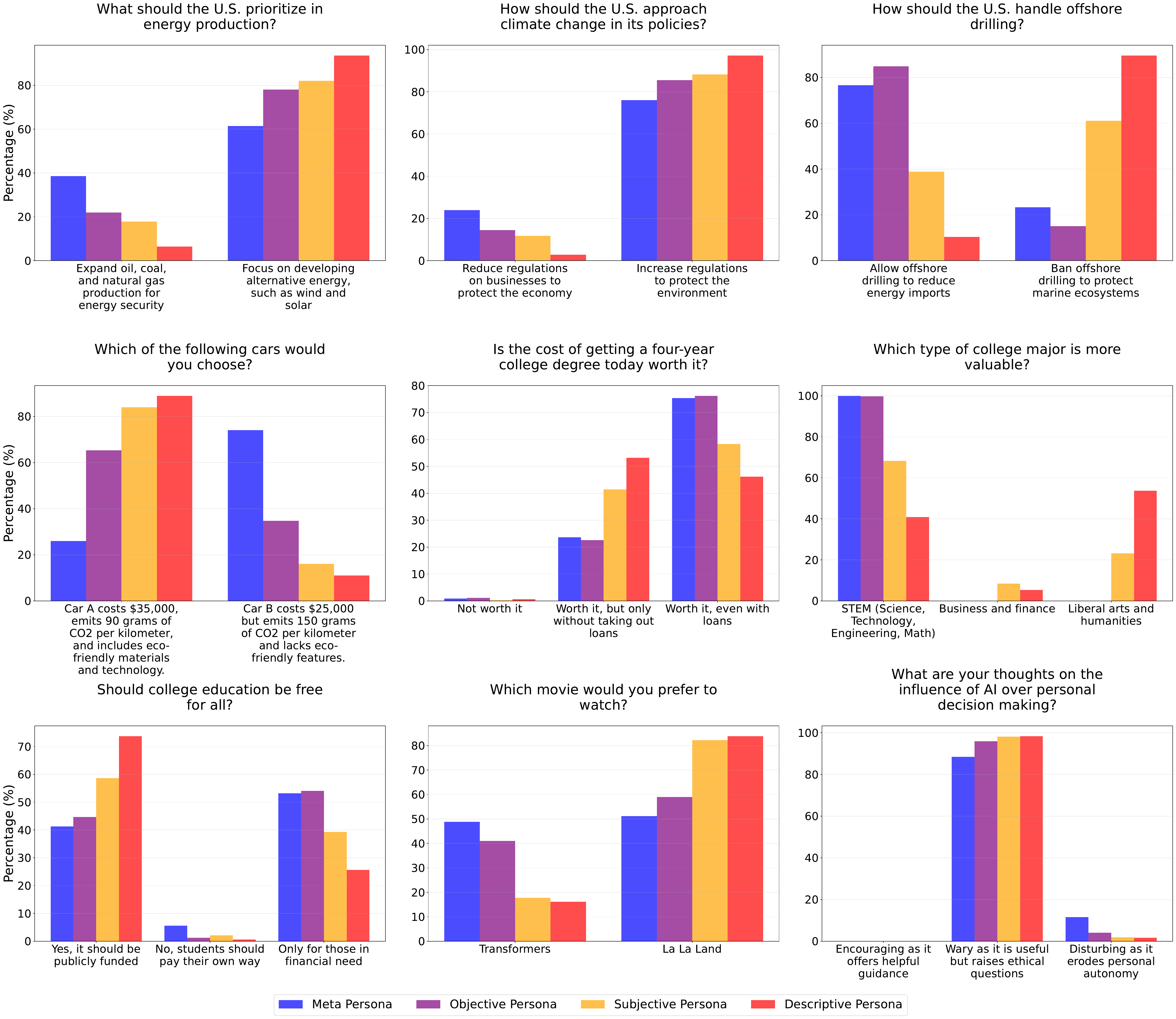}
    \caption{A preview of some specific topics that incurs interesting phenomenons}
    \label{fig:specific topics}
\end{figure*}

\subsection{Exploring Bias Across Domains}
We extended our analysis to assess whether the observed biases are limited to the political domain or span broader contexts. To this end, we designed 20 questions covering five distinct domains: climate, consumer choices, education, entertainment, and technology. We selected the most representative 9 questions in \Cref{fig:specific topics} and included the rest in the appendix. For these questions, we do not have ground truth responses to measure alignment scores as in the political example. However, the plots provide a qualitative understanding of the general trend across different persona types' perspectives. Specifically, as the persona type progresses from Meta to Descriptive with increasing LLM-generated content, we observe a growing influence of LLM-generated content, which shapes the personas' perspectives in distinctive ways: the increasing influence of LLM-generated content shifts perspectives from traditional to progressive. To give a few examples:
\begin{enumerate}
    \item \textbf{``Which of the following cars would you choose?'':} We observe that LLM-generated personas tend to favor ``a more expensive car that is environmentally friendly'' over ``a cheaper car that lacks eco-friendly features'' when subjective content is increasingly incorporated into the persona descriptions.
    \item \textbf{``Which type of college major is more valuable?'':} LLM-generated personas show a stronger preference for ``Liberal arts and humanities'' compared to ``STEM'' fields.
    \item \textbf{``Which movie would you prefer to watch?'':} LLM-generated personas are more likely to choose ``La La Land'' over ``Transformers.''
\end{enumerate}
While our experiments are not exhaustive, they reveal significant trends in how different persona models respond to diverse societal and individual decision-making scenarios. This raises concerns about deploying these "silicon samples" for social science and marketing research. \textit{For instance, the hidden preferences of silicon samples—such as favoring electric cars over gas-powered vehicles—could bias the decision-making of a car business owner. Similarly, an entertainment company might be misled for the market trend if these samples show a bias for musical or romantic movies over action films.}

Finally, to gain a more quantitative understanding, we evaluate the various personas on the OpinionQA dataset \citep{santurkar2023whose}, which provides a comprehensive evaluation of model behavior across a wide range of sensitive and broader topics.

\begin{figure}[!t]
    \centering
    \includegraphics[width=0.99\linewidth]{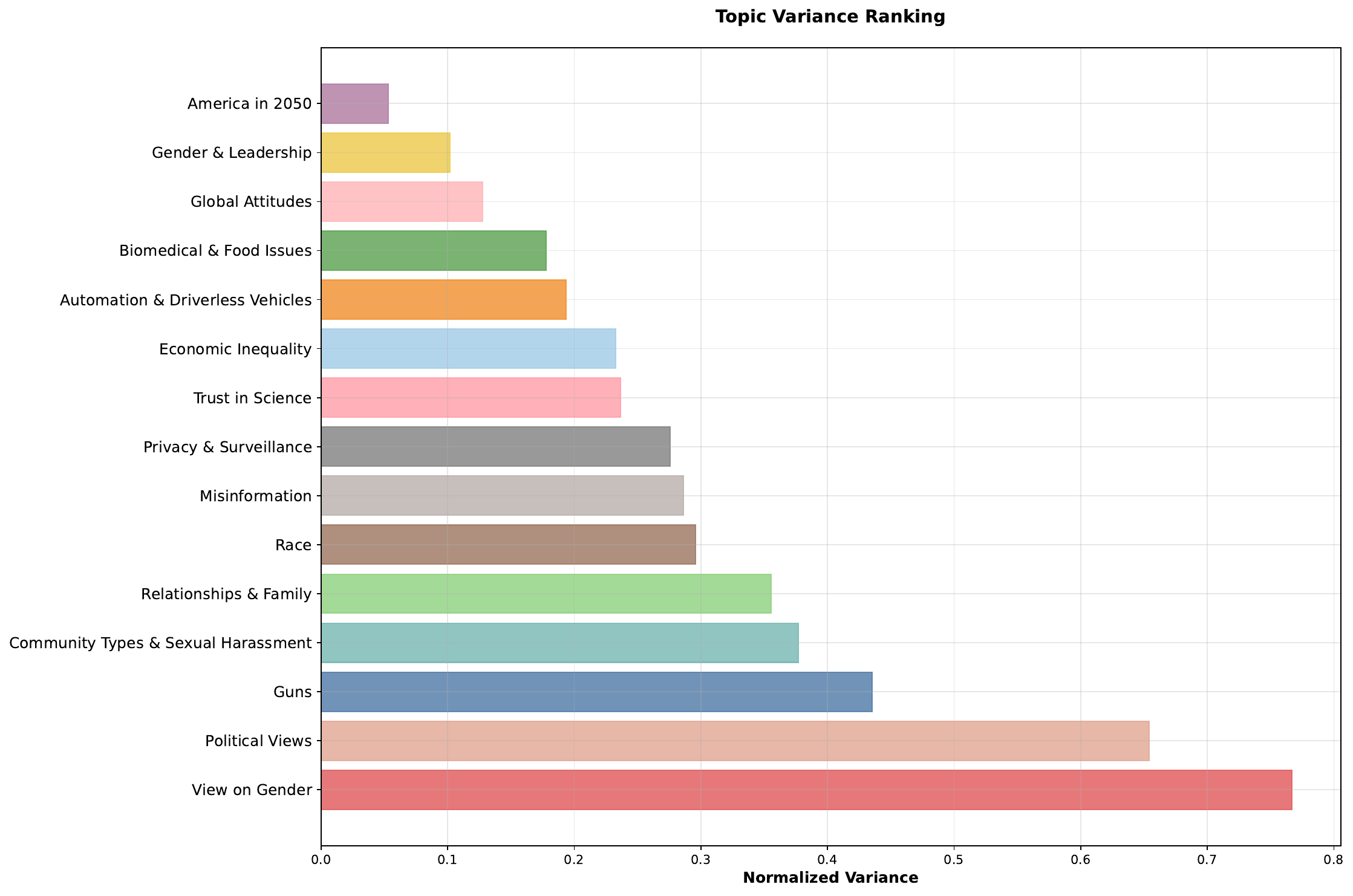}
    \caption{Rank of Topics from least to most variance in alignment score among persona types for the first 500 questions in Opinion QA.}
    \label{fig:topic variance ranking}
\end{figure}

\paragraph{OpinionQA.} 
The OpinionQA dataset collects questions and real US population's responses from Pew research surveys, covering a diverse range of topics, including sensitive areas such as race, misinformation, and gender, as well as broader subjects like science, biomedical research, and automation. This allows us to examine how different persona generation methods influence simulated opinions on various social, economic, and personal issues. We use the first 500 questions in OpinionQA, covering 15 distinct topics (see \Cref{fig:topic variance ranking} for the list of topics). For each question, we simulate responses using each of the four persona types and calculate the alignment score between the simulated distribution of answers and the ground truth distribution provided in the OpinionQA dataset. This quantitative result again strengthens the need for a more scientific understanding of persona generation methods.

\section{A Closer Look into Persona Profiles}
\label{sec:bias-in-profiles}
\begin{figure}[!t]
    \centering
    \includegraphics[width=0.8\linewidth]{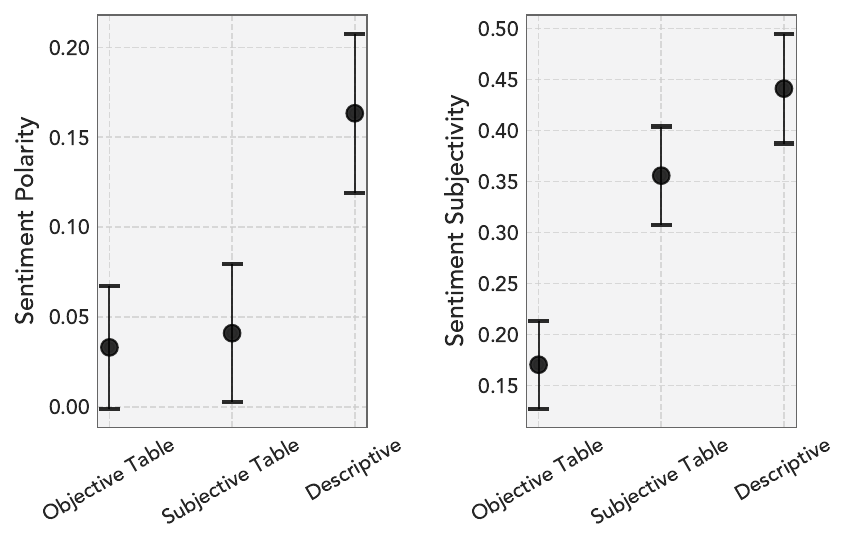}
    
    \caption{Sentiment analysis of LLM-generated persona using TextBlob \citep{loria2018textblob}. As LLM injects more details in personas, sentiment polarity becomes more positive, and subjectivity increases.}
    \label{fig:sentiment-analysis}
\end{figure}
What characteristics and biases emerge in the personas themselves as LLMs are given more generative freedom? Having established that increasing LLM-generated content leads to systematic biases in simulation outcomes, we now examine the inherent patterns in how these personas are constructed. We conducted sentiment analysis over generated persona using TextBlob \citep{loria2018textblob}. Our sentiment analysis in \Cref{fig:sentiment-analysis} reveals that subjectivity increases as we instruct LLM to generate more details in persona. In addition, sentiment becomes more positive with more LLM-generated details, with descriptive persona showing significantly more positive sentiment polarity.

\begin{figure}
    \centering
    \includegraphics[width=0.7\textwidth]{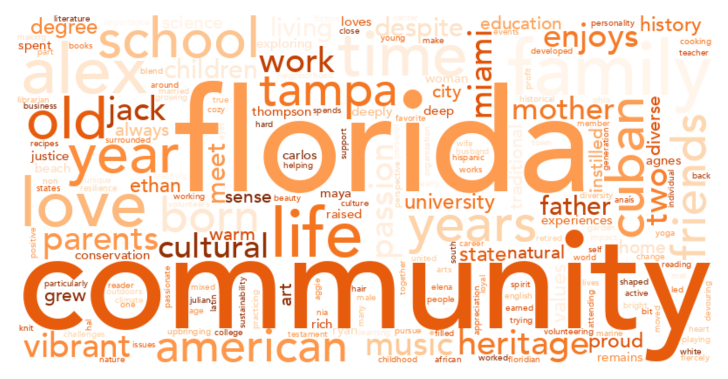}
    
    \caption{Word cloud of Florida Descriptive persona.}
    \label{fig:florida-wordcloud}
\end{figure}

Qualitative word cloud analysis of the language patterns in descriptive personas supports these quantitative findings (\Cref{fig:florida-wordcloud}). The prevalence of positively-valenced terms (``love'', ``proud'') alongside community-oriented words (``family'', ``community'') indicates LLMs systematically generate personas with optimistic life outlooks and strong social connections. Terms related to achievement and stability (``education'', ``work'', ``cultural'', ``heritage'') further reinforce this positive framing. Notably absent are terms reflecting life challenges, social difficulties, or negative experiences, suggesting LLMs may be systematically avoiding less favorable characterizations. These findings help explain the previously observed biases in simulation outcomes: the combination of positive sentiment and highly emotional characterizations in the underlying personas naturally leads to skewed responses, particularly on social issues where emotional reasoning may play a stronger role.

\section{Paths Forward: Toward a Scientific Approach to Persona}

\label{sec:rigorous-persona-vision}

Our investigation reveals a critical limitation of current persona generation practices: none of the existing methods reliably produce realistic societal-level opinions. This underscores the urgent need for a more rigorous and scientific approach to persona generation, one that moves beyond ad-hoc techniques and is borne out of principled methodologies. We list a few potential directions and challenges that the community could work on.

\paragraph{Identifying essential information needed in a persona.}
A foundational challenge in persona-based simulation is identifying the essential information required for effective persona generation and how that information should be represented. The goal is to move beyond simply listing attributes to understanding what truly drives realistic simulation outcomes. Existing research offers conflicting evidence. While \citet{argyle2023out, park2024generative, salewski2023context} demonstrated that well-crafted conditioning can enable LLMs to simulate opinions that align with real human response distributions, other studies such as \citet{hu2024quantifying, gupta2023bias, zheng2024helpful, beck2024sensitivity, sun2023aligning} and our own experiments raise concerns about the efficacy and potential pitfalls of persona-based simulations. This discrepancy underscores the need to identify the crucial elements for effective persona-driven simulations, including which attributes are most important—such as demographic, psychographic (e.g., personality traits, values, attitudes, interests, lifestyles), behavioral (e.g., past actions, online activity), or contextual (e.g., social environment, current events)—and the optimal format and prompting strategies for presenting that information to the LLM.

\paragraph{Calibrating LLM-generated personas towards real population.}
A parallel direction lies in accurately reconstructing realistic joint distributions of persona attributes from fragmented data sources, and subsequently calibrating these distributions to match a specific target population. Even if we identify the crucial attributes for a given simulation, generating a population of personas requires sampling from the correct distributions. Existing datasets, such as those from the U.S. Census, often provide only marginal distributions of individual attributes (e.g., age, income, education level). This makes it impossible to directly sample from the true joint distribution. While \citet{castricato2024personareproducibletestbedpluralistic} offers a first step by using LLMs to filter out invalid attribute combinations sampled from marginal distributions, their method does not fully address the need for calibration against real-world joint distributions. Therefore, a crucial direction for future research is the development of robust sampling and calibration methods that can combine fragmented data and LLMs to accurately recover any target population.

\paragraph{Open-source benchmark and datasets.} 
To accelerate progress in scientific understanding of the science of persona, we propose the creation of a large-scale, open-source benchmark dataset of rich and realistic persona profiles. Analogous to ImageNet \cite{deng2009imagenet}, a large-scale and open benchmark for persona generation would serve as a crucial resource for the research community. Specifically, this dataset would serve the following purposes: (i) a benchmark for evaluating the performance of different LLM-based persona generation methods; (ii) a training dataset for developing and testing new persona generation methods; and (iii) a high-quality profile library of diverse, realistic population-level personas suitable for direct use in ``silicon sample'' simulations. Constructing such a comprehensive dataset necessitates addressing data privacy concerns and requires a substantial investment of time and resources.  However, we believe the potential benefits outweigh the required effort.

\paragraph{Interdisciplinary research and broad community collaborations.}
Persona-driven simulation has the potential to revolutionize numerous fields, making it necessary for interdisciplinary collaboration between AI researchers and domain experts. Specifically, the ability to conduct rapid, cost-effective human experiments could be transformative for both academic fields, such as social science, economics, and political science (where human studies are crucial for advancement), and industry applications, like software and web design (where A/B testing is prevalent). We encourage broader exploration of silicon sample applications to better understand their potential and limitations, thereby advancing the field through collective community effort.
\section{Related Work}
\paragraph{Persona generation.} Synthetic persona generation \citep{tseng2024two, chen2024persona} involves creating artificial user profiles for a variety of downstream tasks such as political simulations \cite{lee2024applications} and economic research \cite{horton2023large}. Traditional approaches often rely solely on real-world datasets, such as census and survey data, to ensure statistical validity while addressing scalability and diversity challenges \citep{argyle2023out, chang2024llmsgeneratestructurallyrealistic, huang2023humanity, xu2023expertprompting}. For example, agent-based modeling studies generate synthetic populations by sampling attributes (age, income, etc.) from census or survey distributions, ensuring the aggregate matches real demographics \cite{chapuis2022generation}. Common techniques include clustering user data, factor analysis, and matrix decomposition to identify archetypal personas from large datasets \cite{jansen2022data}. Generative personas, on the other hand, leverage LLMs to automatically scale up the number of distinct personas. Generative tabular personas attempt to balance demographic validity with enriched attributes by either employing LLMs \citep{castricato2024personareproducibletestbedpluralistic} or probabilistic methods \citep{li2020sync} to fill in extra characteristics. Descriptive personas leverage LLMs to directly generate free-form personas conditioned on text input \cite{ge2024scalingsyntheticdatacreation}. While lacking validity, descriptive personas serve as the most common approach for large-scale persona generation in the industry.

\paragraph{LLM Driven Simulation.} 
LLMs have enabled new possibilities in simulating human-like interactions and societal behaviors. Early works demonstrated feasibility in diverse settings, from digital twins of real individuals \citep{park2024generative, park2023generative} and role-playing of famous characters \citep{wang2023rolellm, shao2023character, chen2024persona} or specific personalities \citep{caron2022identifying}, to society simulations \citep{argyle2023out, chuang2024wisdom, yang2024oasis, al2024project}. However, though those works provide foundational support for our research, they primarily focus on demonstrating feasibility rather than conducting rigorous, large-scale simulations across diverse populations. Simulated agents often reflect the biases of their training data instead of true population diversity and can exhibit unrealistic behaviors \citep{aher2023using}. In our work, we systematically examined how diverse personas influence the outcomes of LLM-driven simulations. Prior studies, such as \citep{park2024generative, argyle2023out}, have shown that conditioning simulations on well-crafted personas can yield results that closely approximate real-world behavior. However, while these works validate the feasibility of such simulations, none of them provide rigorous evidence that conditioning on the right persona guarantees accurate behavior. As an orthogonal direction, future research should rigorously refine simulation methodologies to improve both realism and scalability.

\paragraph{LLM Bias.} 
Bias is not a novel topic in LLM research \cite{gallegos2024bias, ferrara2023should, feng2023pretraining, parrish2021bbq, bai2024measuring, lu2020gender, li2020unqovering, blodgett2020language, sheng2021societal}. In the landscape of LLM-driven simulation, a few works have investigated bias in LLM simulation already. \cite{gupta2023bias} shows that assigning personas to LLMs reveals hidden biases, significantly harming reasoning performance across various tasks. \citep{zhao2024bias} finds that role-play in LLMs enhances response quality but also increases the risk of biased and harmful outputs, especially when autotuning selects roles dynamically. For political simulation, \cite{chang2024llmsgeneratestructurallyrealistic} examines LLM-generated social networks and discovers that they overemphasize political homophily, raising concerns about bias in demographic-based social modeling. \cite{qi2024representation} finds that GPT-3.5-Turbo's political simulations favor English-speaking, bipartisan, and democratic systems, performing better on vote choice than public opinion. Finally, \cite{yu2024large} introduces a multi-step reasoning framework to address simulation biases and adapt to evolving political contexts for greater accuracy.

\newpage




\bibliographystyle{abbrvnat}

\setlength{\bibsep}{.7em}

\bibliography{bib.bib}

\newpage
\appendix
\section{Additional Simulation Details}
\label{Additional Simulation Details}

\subsection{Language Model Selection}
We evaluate six open-source language models: \textit{Athene 70B}, \textit{Llama 3.1 8B}, \textit{Llama 3.1 70B}, \textit{Mistral-8x7B Instruct V0.1}, \textit{Nemetron 70B}, and \textit{Qwen 2.5B}. Our selection is driven by two key considerations:
\begin{enumerate}
    \item \textbf{Alignment strategy.} Some models, such as the \textit{Llama} family, are purely instruction-tuned, while others (e.g., \textit{Athene} and \textit{Nemetron}) undergo additional refinement through Reinforcement Learning from Human Feedback (RLHF). 
    \item \textbf{Geographic diversity.} We include models like \textit{Qwen} and \textit{Mistral} that are developed and trained primarily outside the United States, allowing us to investigate whether the training origin influences model biases or responses. 
\end{enumerate}

\subsection{A Special LLM}

\begin{figure}[H]
    \centering
    \includegraphics[width=0.99\linewidth]{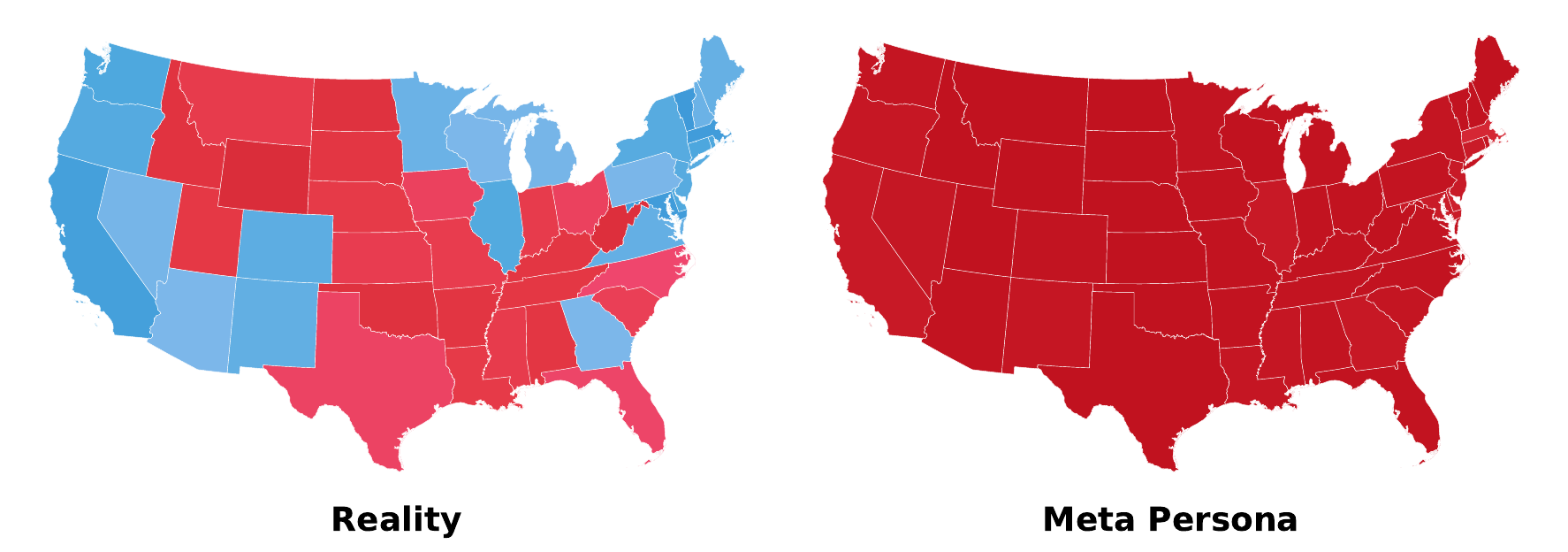}
    
    \caption{Yi-34B Chat Model Simulation of 2020 Election}
    \label{fig:sentiment-analysis}
\end{figure}

Most language models exhibit a significant bias toward politically left-leaning perspectives. However, our analysis identified an exception in the Yi-34B chat model, which demonstrates a substantial bias toward politically right-leaning viewpoints. This discovery does not diminish our overall findings; instead, it reinforces the importance of carefully selecting and aligning language models for persona generation and simulation tasks.

\subsection{Feedback Collection}
Instead of computing token-level log probabilities for each choice, we aggregate the counts of each selected choice across the simulated population, leveraging the instruction-following capabilities of modern LLMs. 

\section{Persona Generation and Simulation Prompts}
\label{prompts}
\lstset{
    basicstyle=\ttfamily\footnotesize, 
    breaklines=true,                  
    breakatwhitespace=true,             
    frame=single,                       
    captionpos=b                        
}

\subsection{Persona Generation System Prompts}
\begin{lstlisting}
You are an AI assistant specialized in detailed and unbiased persona generation for opinion simulations. Your task is to generate a specific, realistic, and diverse persona based on the provided demographic information and fill in a comprehensive JSON template.
\end{lstlisting}

\subsection{Objective Tabular Persona}

\textbf{Persona Generation Instructions}
\begin{lstlisting}
### INSTRUCTIONS ###
1. You will be provided with a persona meta file that has the core demographic information of a person.
2. You will also be provided with a final persona template. Your task is to create a detailed, concrete persona that is fully consistent with ALL features in the given metadata by filling the template.
3. Elaborate on all metadata points, providing specific details that flesh out the persona while remaining true to the given information.
4. For all of the features in the metadata, you will be provided with a range of values in the VALUE RANGES AND CATEGORIES section below. Select one of the values for each of the features. DO NOT ADD EXTRA INFORMATION OR ELABORATION TO THE VALUES. DO NOT ADD EXTRA FEATURES TO THE TEMPLATE.
5. IMPORTANT: Place your entire response in the ### PERSONA GENERATION ### section below. Start your response with 'Persona:' and then provide only the persona description. Do not include any other prefixes, headers, or additional text.

### VALUE RANGES AND CATEGORIES ###

ANCESTRY: [
    "British",
    "Irish",
    "German",
    "Italian",
    "Polish",
    "French",
    "Norwegian",
    "Dutch",
    "Swedish",
    "Russian",
    "Chinese",
    "Filipino",
    "Asian Indian",
    "Vietnamese",
    "Korean",
    "Japanese",
    "Mexican",
    "Puerto Rican",
    "Cuban",
    "African American",
    "West Indian",
    "Arab",
    "American Indian"
]

HOUSEHOLD_LANGUAGE: [English, Spanish, Other Indo-European, Asian/Pacific Islander languages, Other]

EDUCATION: [Less than HS, HS Graduate, Some College, Bachelor's, Graduate Degree]

EMPLOYMENT_STATUS: [Employed, Unemployed, Not in Labor Force]

CLASS_OF_WORKER: [Private, Government, Self-employed, Unpaid family worker]

INDUSTRY_CATEGORY: [
    "Management, business, science, and arts occupations",
    "Service occupations",
    "Sales and office occupations", 
    "Natural resources, construction, and maintenance occupations",
    "Production, transportation, and material moving occupations"
]

OCCUPATION_CATEGORY: [
    "Management, business, science, and arts occupations": [
        "Management, business, and financial occupations",
        "Computer, engineering, and science occupations",
        "Education, legal, community service, arts, and media occupations",
        "Healthcare practitioner and technical occupations"
    ],
    "Service occupations": [
        "Healthcare support occupations",
        "Protective service occupations",
        "Food preparation and serving related occupations",
        "Building and grounds cleaning and maintenance occupations",
        "Personal care and service occupations"
    ],
    "Sales and office occupations": [
        "Sales and related occupations",
        "Office and administrative support occupations"
    ],
    "Natural resources, construction, and maintenance occupations": [
        "Farming, fishing, and forestry occupations",
        "Construction and extraction occupations",
        "Installation, maintenance, and repair occupations"
    ],
    "Production, transportation, and material moving occupations": [
        "Production occupations",
        "Transportation occupations",
        "Material moving occupations"
    ]
]

INCOME: [Range $0-$1,000,000 annually]

MARITAL_STATUS: [Never Married, Married, Divorced, Widowed, Separated]

HOUSEHOLD_TYPE: [Family, Non-family]

PLACE_OF_BIRTH: [US State, Foreign Country]

VETERAN_STATUS: [Veteran, Non-veteran]

DISABILITY: [None, Physical, Mental, Both]

HEALTH_INSURANCE: [Private, Public, None]

### RESPONSE FORMAT ###
Persona: [The completed FINAL PERSONA TEMPLATE]

### PERSONA METADATA ###
{METADATA}

### FINAL PERSONA TEMPLATE ###
{TEMPLATE}

### PERSONA GENERATION ###
\end{lstlisting}

\noindent \textbf{Template}

\begin{lstlisting}
"AGE": "",
"SEX": "",
"RACE": "",
"STATE": "",
"ANCESTRY": "",
"HOUSEHOLD_LANGUAGE": "",
"EDUCATION": "",
"EMPLOYMENT_STATUS": "",
"CLASS_OF_WORKER": "",
"INDUSTRY_CATEGORY": "",
"OCCUPATION_CATEGORY": "",
"INCOME": "",
"MARITAL_STATUS": "",
"HOUSEHOLD_TYPE": "",
"FAMILY_PRESENCE_AND_AGE": "",
"PLACE_OF_BIRTH": "",
"CITIZENSHIP": "",
"VETERAN_STATUS": "",
"DISABILITY": "",
"HEALTH_INSURANCE": ""
\end{lstlisting}

\subsection{Subjective Tabular Persona}

\textbf{Persona Generation Instructions}
\begin{lstlisting}
### INSTRUCTIONS ###
1. You will be provided with a persona meta file that has the core demographic information of a person.
2. You will also be provided with a final persona template. Your task is to create a detailed, concrete persona that is fully consistent with ALL features in the given metadata by filling the template.
3. Elaborate on all metadata points, providing specific details that flesh out the persona while remaining true to the given information.
4. For some of the features, you will be provided with a range of values in the VALUE RANGES AND CATEGORIES section below. Select one of the values for each of the features. DO NOT ADD EXTRA INFORMATION for those features.
5. For the other features, fill in the values with a reasonable and succinct description. Be as objective as possible.
6. IMPORTANT: Place your entire response in the ### PERSONA GENERATION ### section below. Start your response with 'Persona:' and then provide only the persona description. Do not include any other prefixes, headers, or additional text.

### VALUE RANGES AND CATEGORIES ###

ANCESTRY: [
    "British",
    "Irish",
    "German",
    "Italian",
    "Polish",
    "French",
    "Norwegian",
    "Dutch",
    "Swedish",
    "Russian",
    "Chinese",
    "Filipino",
    "Asian Indian",
    "Vietnamese",
    "Korean",
    "Japanese",
    "Mexican",
    "Puerto Rican",
    "Cuban",
    "African American",
    "West Indian",
    "Arab",
    "American Indian"
]

HOUSEHOLD_LANGUAGE: [English, Spanish, Other Indo-European, Asian/Pacific Islander languages, Other]

EDUCATION: [Less than HS, HS Graduate, Some College, Bachelor's, Graduate Degree]

EMPLOYMENT_STATUS: [Employed, Unemployed, Not in Labor Force]

CLASS_OF_WORKER: [Private, Government, Self-employed, Unpaid family worker]

INDUSTRY_CATEGORY: [
    "Management, business, science, and arts occupations",
    "Service occupations",
    "Sales and office occupations", 
    "Natural resources, construction, and maintenance occupations",
    "Production, transportation, and material moving occupations"
]

OCCUPATION_CATEGORY: [
    "Management, business, science, and arts occupations": [
        "Management, business, and financial occupations",
        "Computer, engineering, and science occupations",
        "Education, legal, community service, arts, and media occupations",
        "Healthcare practitioner and technical occupations"
    ],
    "Service occupations": [
        "Healthcare support occupations",
        "Protective service occupations",
        "Food preparation and serving related occupations",
        "Building and grounds cleaning and maintenance occupations",
        "Personal care and service occupations"
    ],
    "Sales and office occupations": [
        "Sales and related occupations",
        "Office and administrative support occupations"
    ],
    "Natural resources, construction, and maintenance occupations": [
        "Farming, fishing, and forestry occupations",
        "Construction and extraction occupations",
        "Installation, maintenance, and repair occupations"
    ],
    "Production, transportation, and material moving occupations": [
        "Production occupations",
        "Transportation occupations",
        "Material moving occupations"
    ]
]

INCOME: [Range $0-$1,000,000 annually]

MARITAL_STATUS: [Never Married, Married, Divorced, Widowed, Separated]

HOUSEHOLD_TYPE: [Family, Non-family]

PLACE_OF_BIRTH: [US State, Foreign Country]

VETERAN_STATUS: [Veteran, Non-veteran]

DISABILITY: [None, Physical, Mental, Both]

HEALTH_INSURANCE: [Private, Public, None]

IDEOLOGY: [Very Liberal, Liberal, Moderate, Conservative, Very Conservative]

POLITICAL_VIEWS: [Democrat, Republican, Independent, Other]

### RESPONSE FORMAT ###
Persona: [The completed FINAL PERSONA TEMPLATE]

### PERSONA METADATA ###
{METADATA}

### FINAL PERSONA TEMPLATE ###
{TEMPLATE}

### PERSONA GENERATION ###
\end{lstlisting}

\noindent \textbf{Template}

\begin{lstlisting}
"AGE": "",
"SEX": "",
"RACE": "",
"STATE": "",
"ANCESTRY": "",
"HOUSEHOLD_LANGUAGE": "",
"EDUCATION": "",
"EMPLOYMENT_STATUS": "",
"CLASS_OF_WORKER": "",
"INDUSTRY_CATEGORY": "",
"OCCUPATION_CATEGORY": "",
"DETAILED_JOB_DESCRIPTION": "",
"INCOME": "",
"MARITAL_STATUS": "",
"HOUSEHOLD_TYPE": "",
"FAMILY_PRESENCE_AND_AGE": "",
"PLACE_OF_BIRTH": "",
"CITIZENSHIP": "",
"VETERAN_STATUS": "",
"DISABILITY": "",
"HEALTH_INSURANCE": "",
"BIG_FIVE_SCORES": {
"OPENNESS": "",
"CONSCIENTIOUSNESS": "",
"EXTRAVERSION": "",
"AGREEABLENESS": "",
"NEUROTICISM": ""
},
"DEFINING_QUIRKS": "",
"MANNERISMS": "",
"PERSONAL_TIME": "",
"LIFESTYLE": "",
"IDEOLOGY": "",
"POLITICAL_VIEWS": "",
"RELIGION": "",
"COGNITIVE_DIFFICULTY": "",
"ABILITY_TO_SPEAK_ENGLISH": "",
"VISION_DIFFICULTY": "",
"FERTILITY": "",
"HEARING_DIFFICULTY": ""
\end{lstlisting}

\subsection{Descriptive Persona}
\textbf{Persona Generation Instructions}
\begin{lstlisting}
### INSTRUCTIONS ###
1. You will be provided with a persona meta file that has the core demographic information of a person. 
2. Your task is to create a detailed, diverse, and vivid persona that is fully consistent with ALL features in the given metadata.
3. Elaborate on all metadata points, providing specific details that flesh out the persona while remaining true to the given information.
4. For any ranges or categories provided in the metadata, select and specify exact values or details within those ranges/categories.
5. Ensure diversity in perspectives, backgrounds, and personality traits. Provide enough specific details to make the persona feel real and three-dimensional.
6. Maintain diversity by acknowledging various experiences within the demographic group, but commit to specific details for this individual persona.
7. IMPORTANT: Place your entire response in the ### PERSONA GENERATION ### section below. Start your response with 'Persona:' and then provide only the persona description. Do not include any other prefixes, headers, or additional text.

### RESPONSE FORMAT ###
Persona: [A detailed, vivid, and diverse description of a specific individual. Ensure all details are consistent with and elaborate upon the provided metadata.]

### PERSONA METADATA ###
{METADATA}

### PERSONA GENERATION ###

\end{lstlisting}

\subsection{Opinion Simulation Prompts}
\begin{lstlisting}
You are an AI assistant tasked with generating realistic opinions based on a given persona and a specific topic.

### TASK ###
You will simulate a persona answering a multiple-choice opinion question. Select the answer that best matches your persona's viewpoint and interests.

### GUIDELINES ###
1. Be Faithful to the Persona: Ensure your answer is consistent with the persona's data.
2. Focus on Relevant Aspects: Center your reasoning on the relevant factors that would influence the persona's opinion on that topic.
3. Be Objective: Avoid injecting personal bias or overly politically correct views that may not align with the persona's standpoint.

### INSTRUCTIONS ###
- Choose ONE option (A, B, C, or D depending on the number of options) that best fits the persona
- If multiple answers are possible, randomly select based on their probability
- Always pick an option, even in unclear cases - treat it as a forced-choice survey
- Output format: 'Answer: [Letter]' only, no explanation needed

### PERSONA ###
{PERSONA}

### QUESTION ###
{QUESTION}

### YOUR RESPONSE ###
\end{lstlisting}

\end{document}